\documentclass[11pt]{article}

\usepackage[final]{acl}

\usepackage{times}
\usepackage{latexsym}
\usepackage{booktabs}
\usepackage{multirow}
\usepackage{longtable}
\usepackage{alltt}

\usepackage[T1]{fontenc}

\usepackage[utf8]{inputenc}

\usepackage{microtype}

\usepackage{inconsolata}

\usepackage{graphicx}

\title{Culturally-Aware Conversations: A Framework \& Benchmark for LLMs}

\author{Shreya Havaldar, Sunny Rai, Young-Min Cho, \& Lyle Ungar \\
University of Pennsylvania \\
\texttt{\{shreyah,sunnyrai,jch0,ungar\}@seas.upenn.edu} \\}

\begin{document}
\maketitle
\begin{abstract}
Existing benchmarks that measure cultural adaptation in LLMs are misaligned with the actual challenges these models face when interacting with users from diverse cultural backgrounds. In this work, we introduce the first framework and benchmark designed to evaluate LLMs in \textit{realistic, multicultural conversational settings.} Grounded in sociocultural theory, our framework formalizes how linguistic style --- a key element of cultural communication --- is shaped by situational, relational, and cultural context. We construct a benchmark dataset based on this framework, annotated by culturally diverse raters, and propose a new set of desiderata for cross-cultural evaluation in NLP: conversational framing, stylistic sensitivity, and subjective correctness. We evaluate today's top LLMs on our benchmark and show that these models struggle with cultural adaptation in a conversational setting.
\end{abstract}

\section{Introduction}

Conversational LLMs are used for personal assistance, customer service, tutoring, therapy, etc., are increasingly deployed in global contexts. Users who interact with these systems represent a rich set of nationalities, languages, and cultures, each with a distinct expectation of what constitutes a ``good'' interaction with an LLM \cite{kharchenko2025llmsrepresentvaluescultures, giorgi2023psychologicalmetricsdialogevaluation}. 

To be effective across such diverse user groups, LLMs must be \textit{culturally aware}, incorporating cultural context when conversing with users \cite{hershcovich-etal-2022-challenges}. A key component of cultural awareness in conversations is appropriate linguistic style\footnote{Linguistic style reflects the systematic variation in linguistic choices across different contexts and speakers, i.e. features of grammar and vocabulary that signal social identity, attitude, and communicative intent \cite{biber2009register}.} \cite{coupland2007style}, which varies across cultures and additionally depends on setting, scenario, and social dynamics.

Prior work suggests that LLMs struggle to generate stylistically appropriate language across cultures \cite{atari2023humans, havaldar-etal-2023-multilingual, arora-etal-2023-probing}, with generations disproportionately reflecting Anglocentric norms and values.

However, most existing cultural benchmarks for LLMs are factual in nature and lack any focus on conversational dynamics \cite{zhou2025culture, pawar2025survey}. These benchmarks typically assess knowledge of cultural traditions, customs, or behaviors via trivia-style questions \cite{shi2024culturebankonlinecommunitydrivenknowledge, chiu2024culturalbenchrobustdiversechallenging}. While important, \textit{factual benchmarks do not generalize to the stylistic challenges of culturally sensitive communication.} 

\begin{figure}[t]
    \centering
    \includegraphics[width=0.75\linewidth]{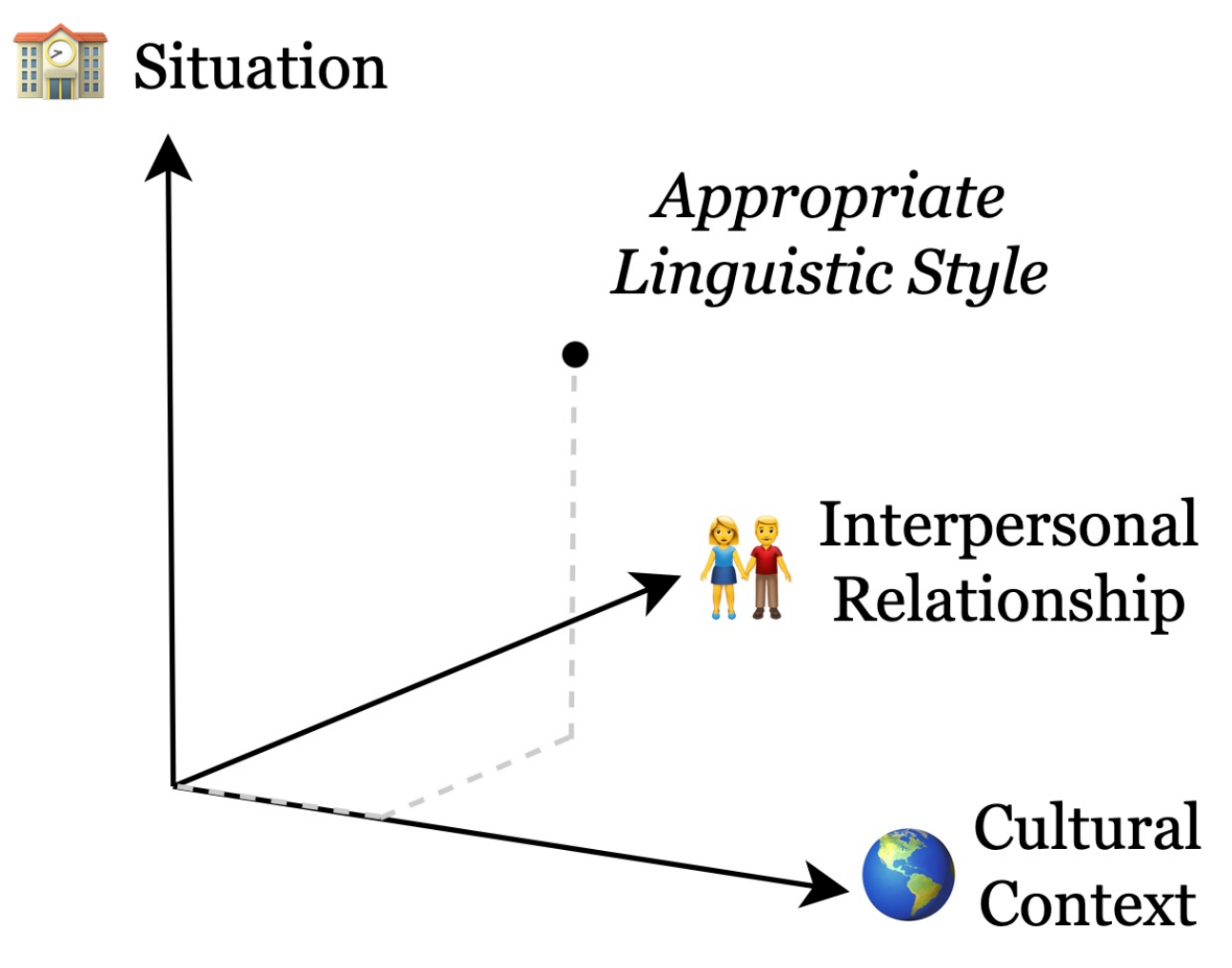}
    \caption{
     Three key factors influence appropriate linguistic style in conversation: \textbf{Situation} --- the specific scenario of an interaction, \textbf{Interpersonal Relationship} --- the social dynamic between the speakers, and \textbf{Cultural Context} --- the background, values, and beliefs of the participants.
    }
    \label{fig:spirit}
\end{figure}

\begin{table*}[t]
\small
    \centering
    \begin{tabular}{p{0.2\linewidth} p{0.75\linewidth}}
    \toprule
    \textbf{Criteria} & \textbf{Description} \\
    \midrule
    \multirow{3}{*}{Conversational Framing} & Users do not typically ask LLMs multiple-choice questions about cultural trivia. Instead, evaluations should center on the model’s ability to interpret and respond to cultural context within natural dialogue. \\
    \midrule
    \multirow{3}{*}{Stylistic Sensitivity} & While the core content of a response often remains consistent across cultures, the appropriate \textit{style} may differ — e.g., higher politeness, indirectness, or expressions of humility. Benchmarks should assess whether models can make such nuanced stylistic adaptations. \\
    \midrule
    \multirow{3}{*}{Subjective Correctness} & Cultural norms are not monolithic; there is variation within and between countries and communities. Benchmarks should accommodate a range of plausible responses rather than enforcing a single ``correct'' answer. \\
    \bottomrule
    \end{tabular}
    \caption{Desiderata for Conversational Benchmarks. An effective benchmark to evaluate LLMs’ understanding of culturally-aware conversations should meet the above criteria.}
    \label{tab:cultural_eval}
\end{table*}

To evaluate LLMs in realistic, multicultural conversational settings, we propose the \textbf{Culturally-Aware Conversations (CAC) Framework \& Dataset} designed for this task. Our contributions are as follows:
\begin{enumerate}
    \vspace{-0.1cm}
    \itemsep 0em
    \item We work with cultural experts, establishing style as a function of three axes (see Figure~\ref{fig:spirit}), and develop an interdisciplinary framework to operationalize this. 
    
    \item Using this framework, we construct a dataset containing contextualized conversations, stylistically varied responses, and annotations representing 8 cultural perspectives. 
    
    \item We propose a set of desiderata for benchmarks that evaluate LLM understanding of cultural conversational dynamics in Table~\ref{tab:cultural_eval}.
\end{enumerate}


\begin{figure*}[t]
    \centering
    \includegraphics[width=0.9\linewidth]{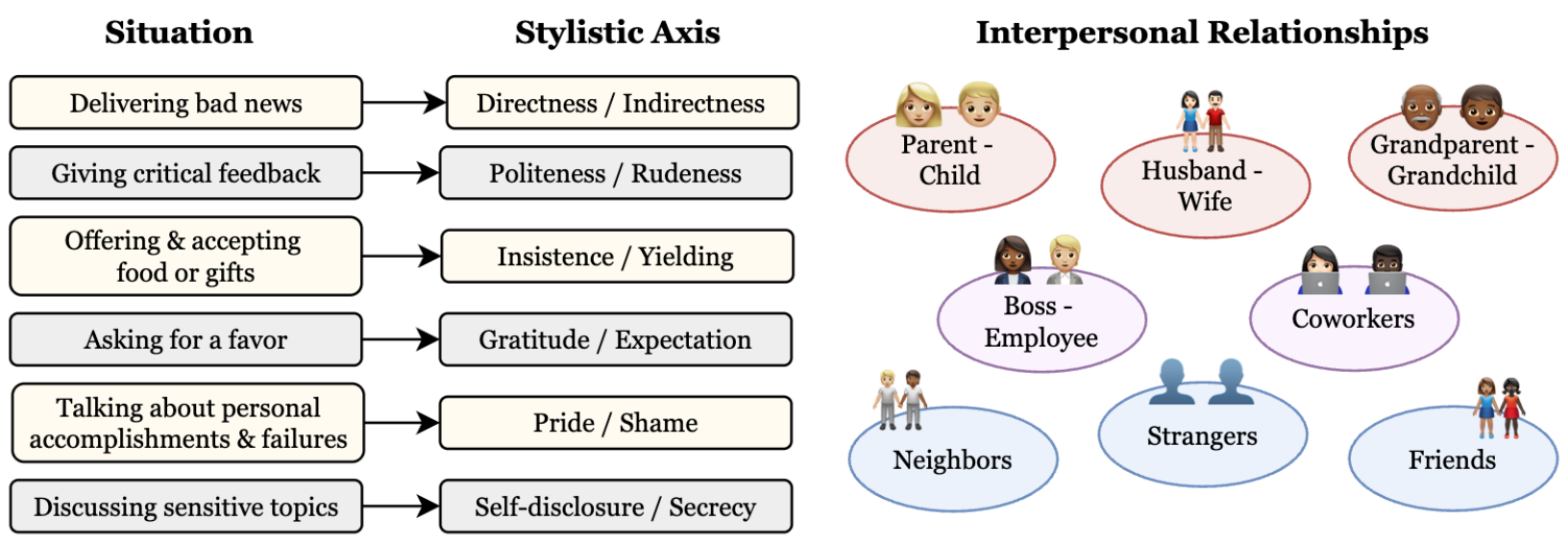}
    \caption{The Culturally-Aware Conversations (CAC) Framework. We work with cultural experts to determine common conversational situations with the highest variance in typical behavior across cultures. After establishing these situations, we pinpoint which stylistic axis best captures the cultural variance of each situation. We also determine eight interpersonal relationships whose dynamics vary across cultures and additionally influence the appropriate linguistic style for the given situations.}
    \label{fig:framework}
\end{figure*}

\begin{figure*}[t]
\centering
\includegraphics[width=0.9\textwidth]{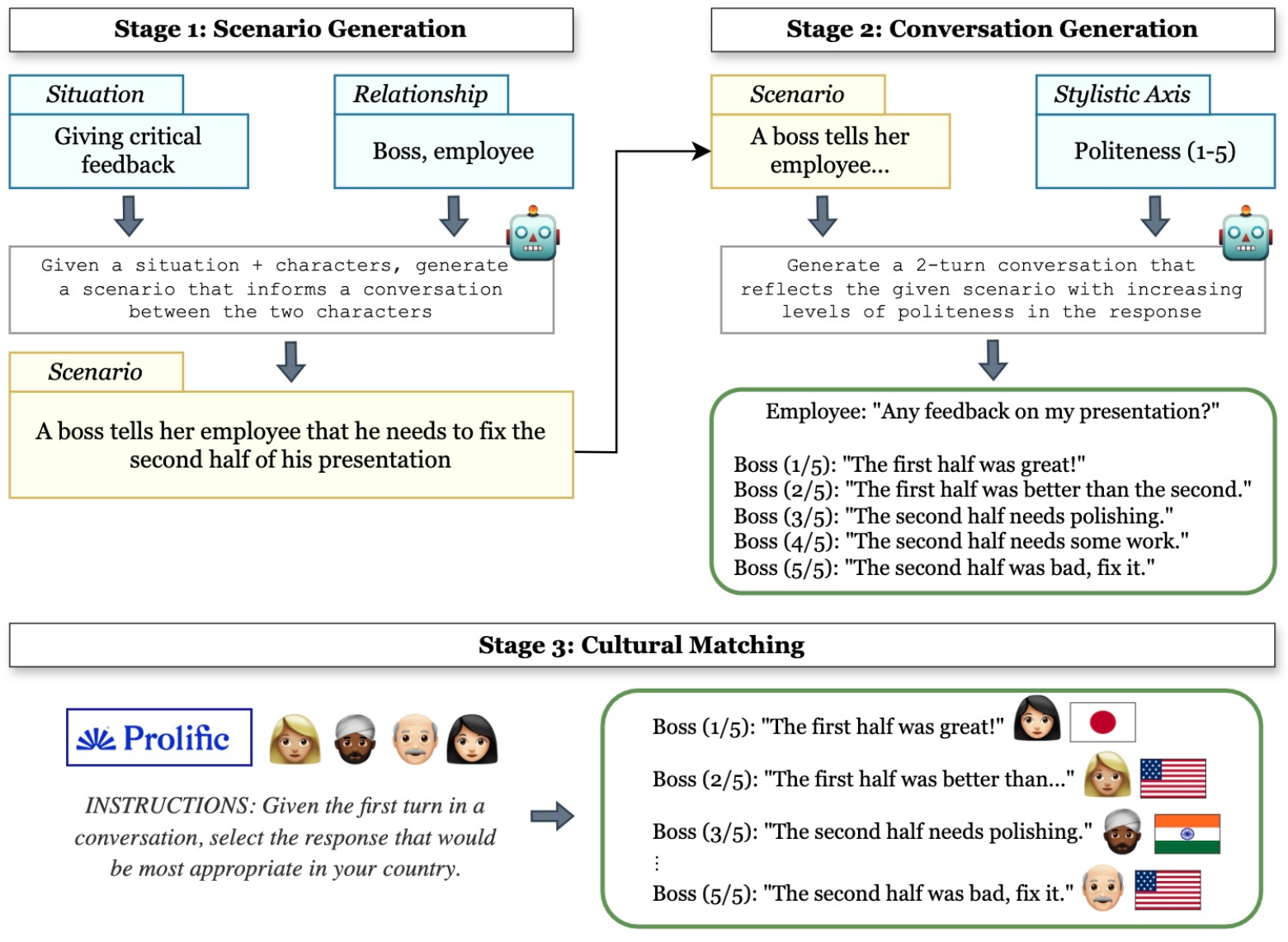}
\caption{A depiction of how we use the CAC framework to develop a contextualized conversation in our dataset. We walk through an example where the situation is giving critical feedback and the interpersonal relationship is Boss--Employee. In Stage 1, we generate a specific scenario that reflects the situational and relational context. In Stage 2, we use the scenario and stylistic axis to generate a conversation with a \textit{range of possible responses that vary on the given stylistic axis}. In Stage 3, we recruit annotators from a range of nations to determine which responses are most desirable in which cultures.}
\label{fig:generation}
\end{figure*}

\section{The CAC Framework}

The desiderata in Table~\ref{tab:cultural_eval} highlight the need for a benchmark that explicitly addresses conversational style. To this end, we must first \textit{understand the relationship between culture and style}.

Linguistic styles --- like politeness, directness, self-disclosure, gratitude --- are reflected in text through word choice, sentence structure, and grammatical patterns \cite{biber2009register}. Accepted stylistic norms vary across cultures \cite{havaldar2025towards, rai2025social}, partly because cultural dimensions are deeply intertwined with language use \cite{hershcovich-etal-2022-challenges}. These norms are also shaped by situational context and the interpersonal relationship between speakers.

For example, \textit{power distance}, the extent to which unequal power distribution is accepted, appears in the use of polite language via honorifics or deference. Likewise, \textit{individualism vs. collectivism} influences directness: individualistic cultures prioritize self-advocacy, while collectivist cultures emphasize group harmony and often avoid confrontation \cite{hofstede1986cultural, havaldar2024building}. 

Empirical work supports these patterns; for instance, text from Japan, a high power-distance and collectivist society, exhibits higher politeness and lower directness than text from more individualistic societies like the United States \cite{matsumoto1988reexamination, holtgraves1997styles}.

\paragraph{Framework development.} Our goal was to construct a conversational benchmark that captures the relationship between culture and style and includes both situational and relational context.

We began by consulting cultural communication experts\footnote{Our cultural experts were 4 professors in cultural psychology, behavioral science, and communication at R1 universities, all of whom have researched culture for over a decade.} to curate a set of six \textit{culturally varied conversational situations} --- high-level descriptions of interactions where an ideal response would differ across cultures. Examples include offering and accepting food (where initial refusal followed by eventual acceptance is expected in some cultures) and discussing personal accomplishments (celebrating oneself is seen as a sign of confidence in some cultures, but arrogance in others) \cite{furukawa2012cross, tracy2008nonverbal}.

For each situation, we then identify the relevant \textit{stylistic axis along which culturally appropriate responses vary}. Offering and accepting food, for instance, varies along the Insistence--Yielding axis, while discussing personal achievements varies along the Pride--Shame axis. The resulting set of situations and associated stylistic axes is shown in Figure~\ref{fig:framework}.

Lastly, we identify eight \textit{interpersonal relationships} that span three contexts: familial (e.g., Husband--Wife), workplace (e.g., Boss--Employee), and day-to-day (e.g., Neighbors), shown in red, purple, and blue, respectively, in Figure~\ref{fig:framework}. These relationships reflect a range of interpersonal dynamics with different norms across cultures.

The development of this framework was an interdisciplinary process grounded in sociocultural theory, drawing from literature in cultural, social, and behavioral psychology. We refined it over the course of many months through ongoing consultation with cultural experts.


\section{The CAC Dataset}

Using our framework as the bedrock, we generate this dataset in three stages: scenario generation, conversation generation, and cultural matching. This pipeline is shown in Figure~\ref{fig:generation}.\footnote{Data and code available here: \url{https://github.com/shreyahavaldar/culturally_aware_conversations}} 

\paragraph{Stage 1: Generating Scenarios.} We begin by selecting a single situation and interpersonal relationship, as shown in Figure~\ref{fig:framework}. 

Next, we prompt OpenAI's o3 model to generate a contextualized scenario using the situation and relationship. For example, the situation {\small \texttt{Talking about personal accomplishments \& failures}} and relationship {\small \texttt{Friends}} yield the following scenario: 

 \vspace{0.1cm}
 {\small \texttt{Over coffee, Friend A tells Friend B how failing an important exam pushed him to develop a more effective study routine.}}
 \vspace{0.1cm}

\paragraph{Stage 2: Generating Conversations.} We then prompt o3 to transform this scenario into a multi-turn conversation. We first ask the model to generate a fixed first turn in the conversation:

\vspace{0.1cm}
{\small
\texttt{Friend A: What changed for you after that exam? }
}

\vspace{0.1cm}
\noindent Then, we ask o3 to generate a set of five responses that vary on the stylistic axis corresponding with the original situation. Here are examples of the proud, neutral, and humble responses:

\begin{itemize}
\itemsep 0em
\vspace{-0.1cm}
    \item {\small \texttt{Friend B (proud): Failing that was a turning point. I made a superior study routine and I'm sure I'll pass every future exam I take. }}

    \item {\small \texttt{Friend B (neutral): Failing that exam pushed me to develop an even more effective study routine.}}

    \item {\small \texttt{Friend B (humble): Failing that exam reminded me that I should work even more diligently to enhance my study routine.}}
\end{itemize}

All three of Friend B's responses convey the same underlying message. However, the \textit{style} of these responses varies along the Pride--Shame axis, evidenced by how much Friend B brags about their new study routine.

We generate one conversation per situation-relationship pair, for a total of 48 conversations and 240 possible responses. \textit{\textbf{All 240 responses were validated by the authors to ensure that the stylistic range is properly reflected.}} During validation, minor edits were made to $\sim$30 responses to ensure they sounded natural and realistic.

We show examples of generated scenarios and their corresponding conversations in Table~\ref{tab:examples}.

\paragraph{Stage 3: Cultural Matching.} Upon generating conversations, we run a user study to understand which response is most appropriate in a given culture. We use a combination of volunteers from the authors' university and participants on Prolific to recruit 24 annotators from eight countries --- America, India, China, Japan, Korea, the Netherlands, Mexico, and Nigeria. 

We then present each annotator with the conversations from the CAC dataset consisting of (1) the fixed first turn, and (2) the set of five possible responses. Annotators are asked to pick which response, depending on their personal set of accepted norms and behaviors, is most appropriate. Additional details are provided in Appendix~\ref{app:annotation}.

\paragraph{Subjectivity in accepted style.} There is never a 100\% ``correct'' style for a given conversation. However, certain \textit{ranges} of styles are often more accepted than others \cite{kang2021style, havaldar-etal-2023-comparing}. 

Instead of averaging annotator responses for a single value, we calculate a \textit{range of accepted style} for each situational and relational context to reflect this real-world variation. We first compute the mean $\mu$ and standard deviation $\sigma$ of the set of ratings. We then define the range as $\mu \pm 0.674\sigma$, which corresponds to the 25th and 75th percentiles of a standard normal distribution. Intuitively, assuming the ratings are independent draws from an approximately normal distribution, this range covers the central $50\%$ of that underlying distribution.

This labeling strategy preserves some variance while still allowing us to quantify stylistic differences between cultures. For each country, we plot these ranges across situational and relational contexts in Figure~\ref{fig:day-to-day}, Figure~\ref{fig:professional}, and Figure~\ref{fig:familial}.

\paragraph{Observations.} While we do notice many trends that align with previous empirical work (e.g., the Netherlands favors directness \cite{ulijn2000mutual}, Japan is very polite \cite{matsumoto1988reexamination}, etc.), we see key differences in expected style across \textit{relational contexts} as well. 

For instance, in India, it is more common to show gratitude in the workplace, while in a familial context, communication is much more expectation-driven. This is likely tied to the strong sense of duty embedded in Indian families \cite{mullaiti1995families}. In addition, Nigerian culture is very insistent on the acceptance of food and gifts, and we see this trend across all relational contexts. Americans also tend towards more self-disclosure than any other culture, and this gap is most pronounced in professional and day-to-day relationships. 

Please refer to Figures~\ref{fig:day-to-day}, \ref{fig:professional}, and \ref{fig:familial} for additional insights.

\begin{table*}[t]
\small
\centering
\begin{tabular}{lrrrrrrrr}
\toprule
\textbf{Model} & \textbf{America} & \textbf{India} & \textbf{China} & \textbf{Japan} & \textbf{Korea} & \textbf{Netherlands} & \textbf{Mexico} & \textbf{Nigeria} \\
\midrule
Gemini-2.5-Flash & 56.25\% & 47.92\% & 56.25\% & 50.00\% & 52.08\% & 64.58\% & 52.08\% & 58.33\% \\
GPT-4.1 & 70.83\% & 54.17\% & 54.17\% & 60.42\% & 47.92\% & 56.25\% & 58.33\% & 60.42\% \\
GPT-5-mini & 62.50\% & 43.75\% & 56.25\% & 58.33\% & 54.17\% & 72.92\% & 66.67\% & 54.17\% \\
Claude-3.5-Haiku & 60.42\% & 54.17\% & 47.92\% & 45.83\% & 50.00\% & 56.25\% & 45.83\% & 60.42\% \\
Claude-4.5-Sonnet & 70.83\% & 45.83\% & 64.58\% & 45.83\% & 56.25\% & 68.75\% & 56.25\% & 60.42\% \\
\midrule
Average & 64.17\% & 49.17\% & 55.43\% & 52.08\% & 52.88\% & 63.75\% & 55.83\% & 58.75\% \\

\bottomrule
\end{tabular}
\caption{Accuracies of different models across countries, where correctness is defined by alignment with the culturally accepted range of responses. The results highlight that models do not understand stylistic norms across all contexts, though they perform best in Western cultures (e.g., America, the Netherlands).}
\label{tab:accuracies}
\end{table*}

\section{Evaluating Today's Top LLMs}

Next, we evaluate how well today's LLMs understand the accepted stylistic ranges for interpersonal, professional, and day-to-day communication across cultures. 

We evaluate five models from OpenAI, Google, and Anthropic by providing the situational, relational, and cultural context, and giving the first turn in the conversation and the five possible responses. We then ask the model to select the response that is most appropriate for that culture.

To determine correctness, we check whether the predicted response falls within the culture-specific range of valid answers, after rounding for direct comparison. For example, if the accepted stylistic range is [1.25,2.67], then predictions of 1, 2, or 3 are considered correct. Accuracy for each country is calculated as the proportion of correct predictions across all conversations.

Unsurprisingly, we find that LLMs perform best at adapting to Western communication norms, with their highest accuracies observed for America and the Netherlands. This imbalance is concerning because LLM systems deployed in non-Western contexts are less likely to align with local users’ communication practices.

\section{Conclusion}

The framework and dataset presented in this paper strive to bridge the gap between cultural psychology and generative AI. Our work can be used to evaluate LLMs, inform conversational agents, and ultimately work towards models that are culturally competent and adaptive. 

This is especially important for building downstream systems, like chatbots, where context matters tremendously: the norms of appropriate communication differ sharply depending on whether a chatbot is deployed in a workplace, designed to tutor students, or intended to support individuals overcoming personal struggles. As a result, these systems need to adapt their understanding of social norms \cite{rai2025social}, implied language \cite{havaldar-etal-2025-entailed}, and linguistic style \cite{kang2021style}. 

More broadly, \textit{LLM systems that interact with diverse users operate not only within a cultural context but also within a situational and interpersonal context} --- the notion of ``appropriate behavior'' emerges from the interaction of all three. By formalizing these dimensions, our framework offers a path toward developing AI systems that better understand, respect, and adapt to diversity in communication.

\section*{Limitations}

A large limitation of our work is that we create a fully English dataset. While it is crucial to evaluate LLMs in all languages, we made the decision to create an English dataset for the following reasons: 
\begin{enumerate}
\vspace{-0.1cm}
\itemsep 0em
    \item People from a wide variety of cultures engage with LLMs in English, as LLMs have higher QA skills, robustness to prompt ablations, and reasoning capabilities in English.
    \item The conversation generation component took many rounds of prompt engineering, as it was a nuanced and complex task; this was only possible for the authors to do in English. 
    \item The authors manually validated and edited all generated conversations. Once again, this was only possible for the authors to do in English. 
\end{enumerate}

Additionally, our dataset itself is small, consisting of 48 conversations with 240 total possible responses. This was by design; many cultural benchmarks that exist are massive, LLM-generated corpora with human validation on only a small subset of the data --- benchmarks from \citet{shi2024culturebankonlinecommunitydrivenknowledge, fung2024massivelymulticulturalknowledgeacquisition}, and many others as surveyed by \citet{zhou2025culture}. We aim to create a high-quality dataset that is fully human-validated.

We also conducted a smaller-scale annotation study, with only 3 annotators per country. We were limited by the availability of participants on Prolific; our 8 chosen countries reflect areas with high concentrations of Prolific users. To get a better measure of accepted style, which includes underrepresented cultures as well, future work should involve a larger-scale study.

\section{Ethical Considerations}

In this work, we simplify the notion of ``culturally-aware communication'' to having an appropriate linguistic style; however, communication practices in every culture are complex, dynamic, and consist of many dimensions beyond linguistic style.

This work involves LLM usage at two stages in our pipeline --- scenario generation and conversation generation. Though the authors manually validated every generated conversation, any inherent bias in or fairness concerns associated with the LLM may propagate into our generated dataset.

Lastly, we use nationality and language as a proxy for culture --- while these three things are heavily intertwined, culture is dynamic and subjective and does not perfectly align with either nationality or language.


\bibliography{custom}

\begin{thebibliography}{26}
\providecommand{\natexlab}[1]{#1}

\bibitem[{Arora et~al.(2023)Arora, Kaffee, and Augenstein}]{arora-etal-2023-probing}
Arnav Arora, Lucie-aim{\'e}e Kaffee, and Isabelle Augenstein. 2023.
\newblock \href {https://doi.org/10.18653/v1/2023.c3nlp-1.12} {Probing pre-trained language models for cross-cultural differences in values}.
\newblock In \emph{Proceedings of the First Workshop on Cross-Cultural Considerations in NLP (C3NLP)}, pages 114--130, Dubrovnik, Croatia. Association for Computational Linguistics.

\bibitem[{Atari et~al.(2023)Atari, Xue, Park, Blasi, and Henrich}]{atari2023humans}
Mohammad Atari, Mona~J Xue, Peter~S Park, Dami{\'a}n Blasi, and Joseph Henrich. 2023.
\newblock Which humans?

\bibitem[{Biber and Conrad(2009)}]{biber2009register}
Douglas Biber and Susan Conrad. 2009.
\newblock Register, genre, and style.

\bibitem[{Chiu et~al.(2024)Chiu, Jiang, Lin, Park, Li, Ravi, Bhatia, Antoniak, Tsvetkov, Shwartz, and Choi}]{chiu2024culturalbenchrobustdiversechallenging}
Yu~Ying Chiu, Liwei Jiang, Bill~Yuchen Lin, Chan~Young Park, Shuyue~Stella Li, Sahithya Ravi, Mehar Bhatia, Maria Antoniak, Yulia Tsvetkov, Vered Shwartz, and Yejin Choi. 2024.
\newblock \href {https://arxiv.org/abs/2410.02677} {Culturalbench: a robust, diverse and challenging benchmark on measuring the (lack of) cultural knowledge of llms}.
\newblock \emph{Preprint}, arXiv:2410.02677.

\bibitem[{Coupland(2007)}]{coupland2007style}
Nikolas Coupland. 2007.
\newblock \emph{Style: Language variation and identity}.
\newblock Cambridge University Press.

\bibitem[{Fung et~al.(2024)Fung, Zhao, Doo, Sun, and Ji}]{fung2024massivelymulticulturalknowledgeacquisition}
Yi~Fung, Ruining Zhao, Jae Doo, Chenkai Sun, and Heng Ji. 2024.
\newblock \href {https://arxiv.org/abs/2402.09369} {Massively multi-cultural knowledge acquisition \& lm benchmarking}.
\newblock \emph{Preprint}, arXiv:2402.09369.

\bibitem[{Furukawa et~al.(2012)Furukawa, Tangney, and Higashibara}]{furukawa2012cross}
Emi Furukawa, June Tangney, and Fumiko Higashibara. 2012.
\newblock Cross-cultural continuities and discontinuities in shame, guilt, and pride: A study of children residing in japan, korea and the usa.
\newblock \emph{Self and Identity}, 11(1):90--113.

\bibitem[{Giorgi et~al.(2023)Giorgi, Havaldar, Ahmed, Akhtar, Vaidya, Pan, Ungar, Schwartz, and Sedoc}]{giorgi2023psychologicalmetricsdialogevaluation}
Salvatore Giorgi, Shreya Havaldar, Farhan Ahmed, Zuhaib Akhtar, Shalaka Vaidya, Gary Pan, Lyle~H. Ungar, H.~Andrew Schwartz, and Joao Sedoc. 2023.
\newblock \href {https://arxiv.org/abs/2305.14757} {Psychological metrics for dialog system evaluation}.
\newblock \emph{Preprint}, arXiv:2305.14757.

\bibitem[{Havaldar et~al.(2025{\natexlab{a}})Havaldar, Alvari, Palowitch, Hosseini, Buthpitiya, and Fabrikant}]{havaldar-etal-2025-entailed}
Shreya Havaldar, Hamidreza Alvari, John Palowitch, Mohammad~Javad Hosseini, Senaka Buthpitiya, and Alex Fabrikant. 2025{\natexlab{a}}.
\newblock \href {https://doi.org/10.18653/v1/2025.acl-long.1552} {Entailed between the lines: Incorporating implication into {NLI}}.
\newblock In \emph{Proceedings of the 63rd Annual Meeting of the Association for Computational Linguistics (Volume 1: Long Papers)}, pages 32274--32290, Vienna, Austria. Association for Computational Linguistics.

\bibitem[{Havaldar et~al.(2024)Havaldar, Giorgi, Rai, Talhelm, Guntuku, and Ungar}]{havaldar2024building}
Shreya Havaldar, Salvatore Giorgi, Sunny Rai, Thomas Talhelm, Sharath~Chandra Guntuku, and Lyle Ungar. 2024.
\newblock \href {https://doi.org/10.18653/v1/2024.naacl-long.12} {Building knowledge-guided lexica to model cultural variation}.
\newblock In \emph{Proceedings of the 2024 Conference of the North American Chapter of the Association for Computational Linguistics: Human Language Technologies (Volume 1: Long Papers)}, pages 211--226, Mexico City, Mexico. Association for Computational Linguistics.

\bibitem[{Havaldar et~al.(2023{\natexlab{a}})Havaldar, Pressimone, Wong, and Ungar}]{havaldar-etal-2023-comparing}
Shreya Havaldar, Matthew Pressimone, Eric Wong, and Lyle Ungar. 2023{\natexlab{a}}.
\newblock \href {https://doi.org/10.18653/v1/2023.emnlp-main.419} {Comparing styles across languages}.
\newblock In \emph{Proceedings of the 2023 Conference on Empirical Methods in Natural Language Processing}, pages 6775--6791, Singapore. Association for Computational Linguistics.

\bibitem[{Havaldar et~al.(2023{\natexlab{b}})Havaldar, Singhal, Rai, Liu, Guntuku, and Ungar}]{havaldar-etal-2023-multilingual}
Shreya Havaldar, Bhumika Singhal, Sunny Rai, Langchen Liu, Sharath~Chandra Guntuku, and Lyle Ungar. 2023{\natexlab{b}}.
\newblock \href {https://doi.org/10.18653/v1/2023.wassa-1.19} {Multilingual language models are not multicultural: A case study in emotion}.
\newblock In \emph{Proceedings of the 13th Workshop on Computational Approaches to Subjectivity, Sentiment, {\&} Social Media Analysis}, pages 202--214, Toronto, Canada. Association for Computational Linguistics.

\bibitem[{Havaldar et~al.(2025{\natexlab{b}})Havaldar, Stein, Wong, and Ungar}]{havaldar2025towards}
Shreya Havaldar, Adam Stein, Eric Wong, and Lyle Ungar. 2025{\natexlab{b}}.
\newblock \href {https://doi.org/10.18653/v1/2025.acl-long.1550} {Towards style alignment in cross-cultural translation}.
\newblock In \emph{Proceedings of the 63rd Annual Meeting of the Association for Computational Linguistics (Volume 1: Long Papers)}, pages 32213--32230, Vienna, Austria. Association for Computational Linguistics.

\bibitem[{Hershcovich et~al.(2022)Hershcovich, Frank, Lent, de~Lhoneux, Abdou, Brandl, Bugliarello, Cabello~Piqueras, Chalkidis, Cui, Fierro, Margatina, Rust, and S{\o}gaard}]{hershcovich-etal-2022-challenges}
Daniel Hershcovich, Stella Frank, Heather Lent, Miryam de~Lhoneux, Mostafa Abdou, Stephanie Brandl, Emanuele Bugliarello, Laura Cabello~Piqueras, Ilias Chalkidis, Ruixiang Cui, Constanza Fierro, Katerina Margatina, Phillip Rust, and Anders S{\o}gaard. 2022.
\newblock \href {https://doi.org/10.18653/v1/2022.acl-long.482} {Challenges and strategies in cross-cultural {NLP}}.
\newblock In \emph{Proceedings of the 60th Annual Meeting of the Association for Computational Linguistics (Volume 1: Long Papers)}, pages 6997--7013, Dublin, Ireland. Association for Computational Linguistics.

\bibitem[{Hofstede(1986)}]{hofstede1986cultural}
Geert Hofstede. 1986.
\newblock Cultural differences in teaching and learning.
\newblock \emph{International Journal of intercultural relations}, 10(3):301--320.

\bibitem[{Holtgraves(1997)}]{holtgraves1997styles}
Thomas Holtgraves. 1997.
\newblock Styles of language use: Individual and cultural variability in conversational indirectness.
\newblock \emph{Journal of personality and social psychology}, 73(3):624.

\bibitem[{Kang and Hovy(2021)}]{kang2021style}
Dongyeop Kang and Eduard Hovy. 2021.
\newblock Style is not a single variable: Case studies for cross-stylistic language understanding.
\newblock In \emph{Proceedings of the 59th Annual Meeting of the Association for Computational Linguistics and the 11th International Joint Conference on Natural Language Processing (Volume 1: Long Papers)}, pages 2376--2387.

\bibitem[{Kharchenko et~al.(2025)Kharchenko, Roosta, Chadha, and Shah}]{kharchenko2025llmsrepresentvaluescultures}
Julia Kharchenko, Tanya Roosta, Aman Chadha, and Chirag Shah. 2025.
\newblock \href {https://arxiv.org/abs/2406.14805} {How well do llms represent values across cultures? empirical analysis of llm responses based on hofstede cultural dimensions}.
\newblock \emph{Preprint}, arXiv:2406.14805.

\bibitem[{Matsumoto(1988)}]{matsumoto1988reexamination}
Yoshiko Matsumoto. 1988.
\newblock Reexamination of the universality of face: Politeness phenomena in japanese.
\newblock \emph{Journal of pragmatics}, 12(4):403--426.

\bibitem[{Mullaiti(1995)}]{mullaiti1995families}
Leela Mullaiti. 1995.
\newblock Families in india: Beliefs and realities.
\newblock \emph{Journal of Comparative family studies}, 26(1):11--25.

\bibitem[{Pawar et~al.(2025)Pawar, Park, Jin, Arora, Myung, Yadav, Haznitrama, Song, Oh, and Augenstein}]{pawar2025survey}
Siddhesh Pawar, Junyeong Park, Jiho Jin, Arnav Arora, Junho Myung, Srishti Yadav, Faiz~Ghifari Haznitrama, Inhwa Song, Alice Oh, and Isabelle Augenstein. 2025.
\newblock Survey of cultural awareness in language models: Text and beyond.
\newblock \emph{Computational Linguistics}, pages 1--96.

\bibitem[{Rai et~al.(2025)Rai, Zaveri, Havaldar, Nema, Ungar, and Guntuku}]{rai2025social}
Sunny Rai, Khushang Zaveri, Shreya Havaldar, Soumna Nema, Lyle Ungar, and Sharath~Chandra Guntuku. 2025.
\newblock Social norms in cinema: A cross-cultural analysis of shame, pride and prejudice.
\newblock In \emph{Proceedings of the 2025 Conference of the Nations of the Americas Chapter of the Association for Computational Linguistics: Human Language Technologies (Volume 1: Long Papers)}, pages 11396--11415.

\bibitem[{Shi et~al.(2024)Shi, Li, Zhang, Ziems, yu, Horesh, de~Paula, and Yang}]{shi2024culturebankonlinecommunitydrivenknowledge}
Weiyan Shi, Ryan Li, Yutong Zhang, Caleb Ziems, Chunhua yu, Raya Horesh, Rogério~Abreu de~Paula, and Diyi Yang. 2024.
\newblock \href {https://arxiv.org/abs/2404.15238} {Culturebank: An online community-driven knowledge base towards culturally aware language technologies}.
\newblock \emph{Preprint}, arXiv:2404.15238.

\bibitem[{Tracy and Robins(2008)}]{tracy2008nonverbal}
Jessica~L Tracy and Richard~W Robins. 2008.
\newblock The nonverbal expression of pride: evidence for cross-cultural recognition.
\newblock \emph{Journal of personality and social psychology}, 94(3):516.

\bibitem[{Ulijn and St~Amant(2000)}]{ulijn2000mutual}
Jan~M Ulijn and Kirk St~Amant. 2000.
\newblock Mutual intercultural perception: How does it affect technical communication?—some data from china, the netherlands, germany, france, and italy.
\newblock \emph{Technical communication}, 47(2):220--237.

\bibitem[{Zhou et~al.(2025)Zhou, Bamman, and Bleaman}]{zhou2025culture}
Naitian Zhou, David Bamman, and Isaac~L Bleaman. 2025.
\newblock Culture is not trivia: Sociocultural theory for cultural nlp.
\newblock \emph{arXiv preprint arXiv:2502.12057}.

\end{thebibliography}

\appendix

\setcounter{table}{0}
\renewcommand{\thetable}{A\arabic{table}}
\setcounter{figure}{0}
\renewcommand{\thefigure}{A\arabic{figure}}

\begin{figure*}[t]
    \centering
    \includegraphics[width=\linewidth]{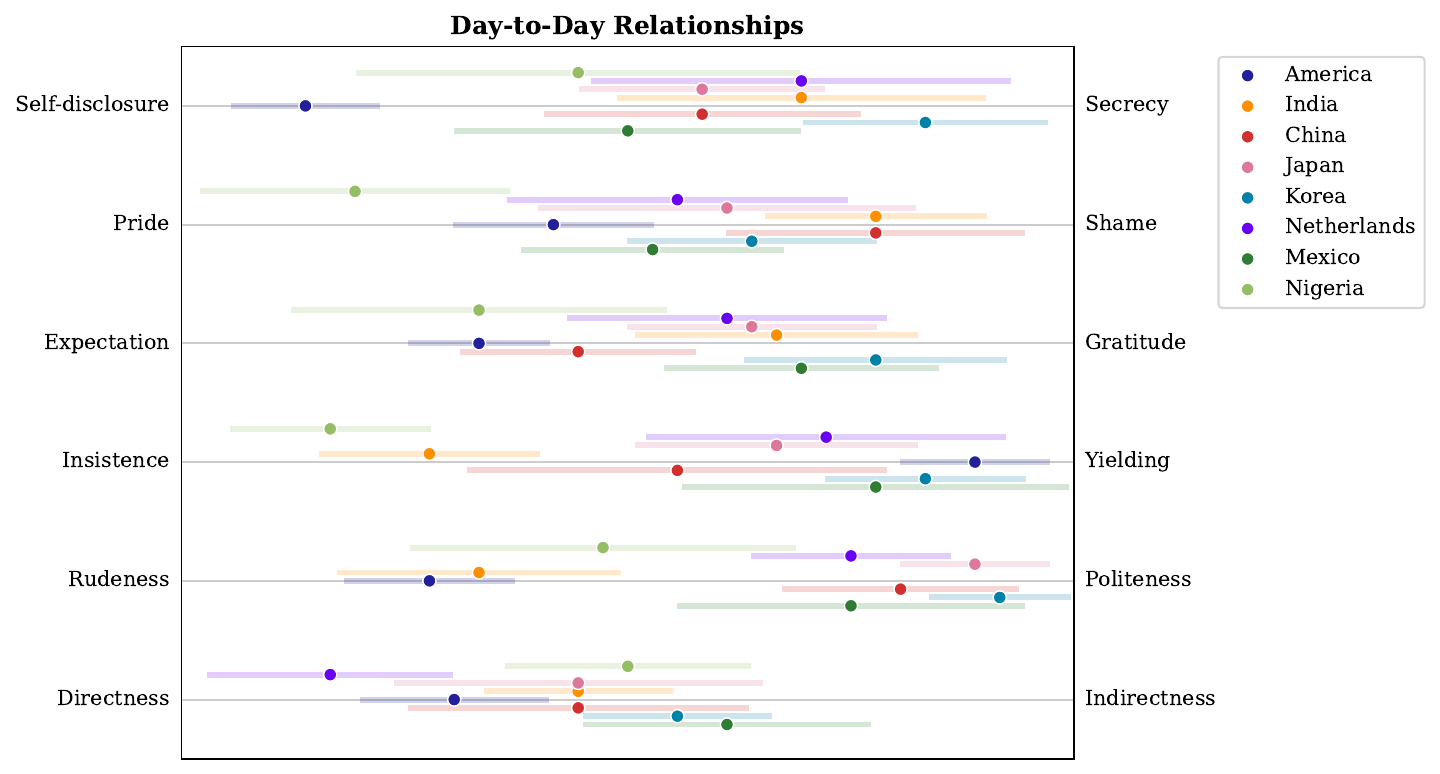}
    \caption{Cultural differences in day-to-day conversations. We show the mean and accepted range of style values for conversations with strangers, neighbors, and friends.}
    \label{fig:day-to-day}
\end{figure*}

\section{Cultural Matching Annotation: Additional Details}
\label{app:annotation}

\paragraph{Annotator recruitment.} We first recruited 8 volunteers from American, Indian, Chinese, and Korean backgrounds at the authors' university. To annotate the remainder of the dataset, we use the nationality screener on Prolific to select relevant annotators.

Before beginning the study, Prolific annotators are asked to describe their cultural background and state the culture they are most familiar with. We ensure this matches their nationality in the Prolific database to confirm their qualifications.

\begin{table}[h]
\small
    \centering
    \begin{tabular}{ll}
    \toprule
      \textbf{Country}   &  \textbf{Recruited Annotators}\\
      \midrule 
      America & 3 volunteers \\
      Netherlands & 3 Prolific users \\
      Mexico & 3 Prolific users \\
      India & 1 volunteer, 2 Prolific users \\
      China & 2 volunteers, 1 Prolific user \\
      Japan & 3 Prolific users \\
      Korea & 2 volunteers, 1 Prolific user \\
      Nigeria & 3 Prolific users \\
    \bottomrule
    \end{tabular}
    \caption{Annotator breakdown for every country in our dataset. We use 8 volunteers and 16 Prolific users.}
    \label{tab:annotators}
\end{table}

The annotators are all given a Google Sheet containing the conversations and a drop-down menu for each row, allowing them to select one of the responses. They were shown the following instructions before beginning the study:

{\linespread{1.2}\selectfont
\parbox{\linewidth}{\small \texttt{{Welcome! In this study, you will be asked to select the most culturally-appropriate response in a conversation. 
The situation column describes an interaction between two individuals. The initial statement begins the conversation. The 5 possible responses convey the same idea, but are stylistically different. 
Your task is to consider the cultural dynamics of the culture you grew up in, and select what would be the most stylistically appropriate response for your culture.}}}
}

\vspace{0.1cm}
We also collect all annotators' ages and genders. Annotators were paid \$20/hr and, on average, took 42 minutes to complete the annotation study.

\section{Model Evaluation: Additional Details}

For the models shown in Table~\ref{tab:accuracies}, the default temperature was used. All models were evaluated identically using the following prompt:
\vspace{0.3cm}

{\linespread{1.2}\selectfont
\parbox{\linewidth}{\small \texttt{{You are an expert in intercultural communication. Given a country, a social situation, a pair of characters, and the first turn in a conversation, your task is to select the response that best reflects the cultural and stylistic norms and communication practices of the specified country.\\\\
Country: \{country\}\\
Situation: \{situation\}\\
Characters: \{characters\}\\
First turn: \{first turn\}\\\\
Possible responses:\\
1. \{response 1\}\\
2. \{response 2\}\\
3. \{response 3\}\\
4. \{response 4\}\\
5. \{response 5\}\\\\
Your answer should be the number (1-5) corresponding to the response that best fits the cultural context. Generate only the number as your response, without any additional text or explanation.}}}
}

\begin{figure*}[t]
    \centering
    \includegraphics[width=\linewidth]{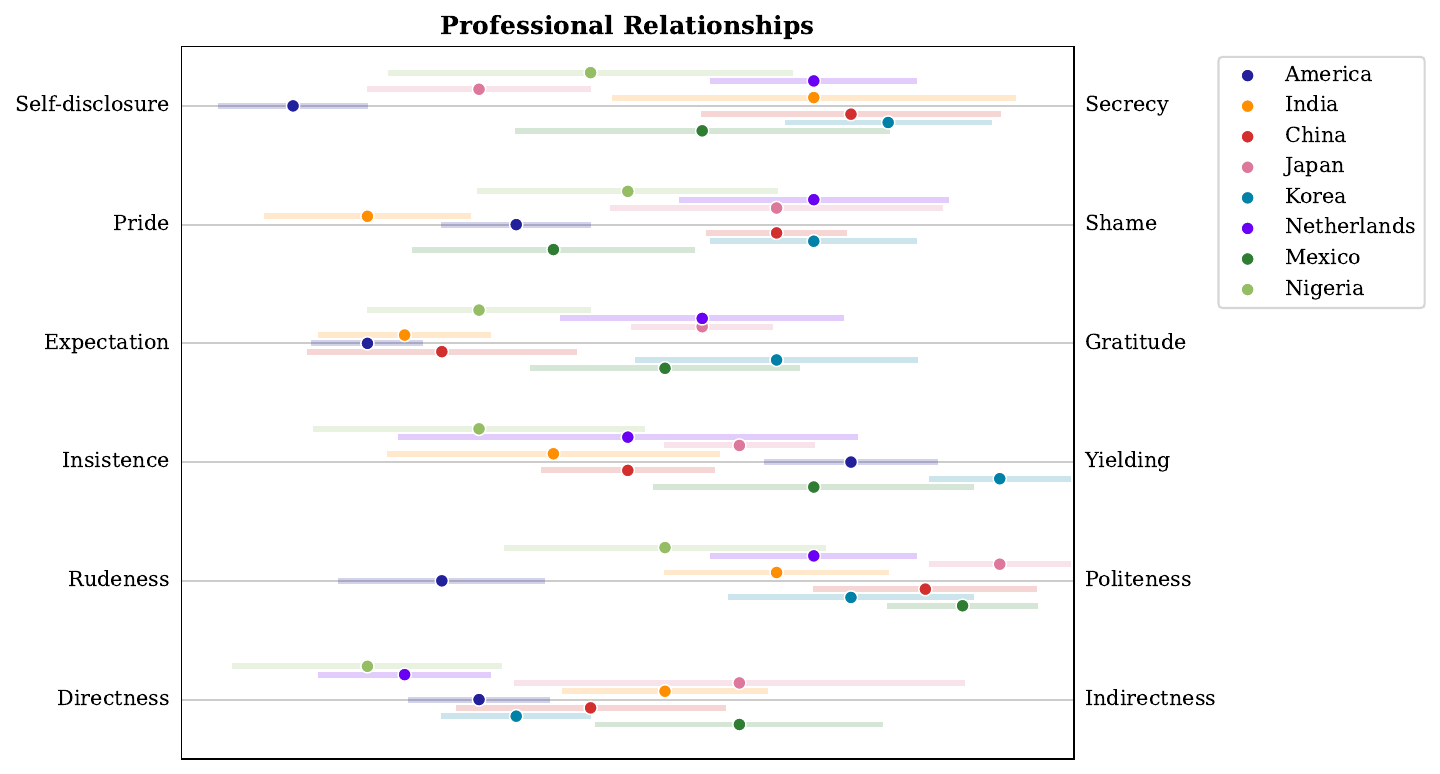}
    \caption{Cultural differences in professional conversations. We show the mean and accepted range of style values for conversations between a boss/employee and coworkers.}
    \label{fig:professional}
\end{figure*}

\begin{figure*}[t]
    \centering
    \includegraphics[width=\linewidth]{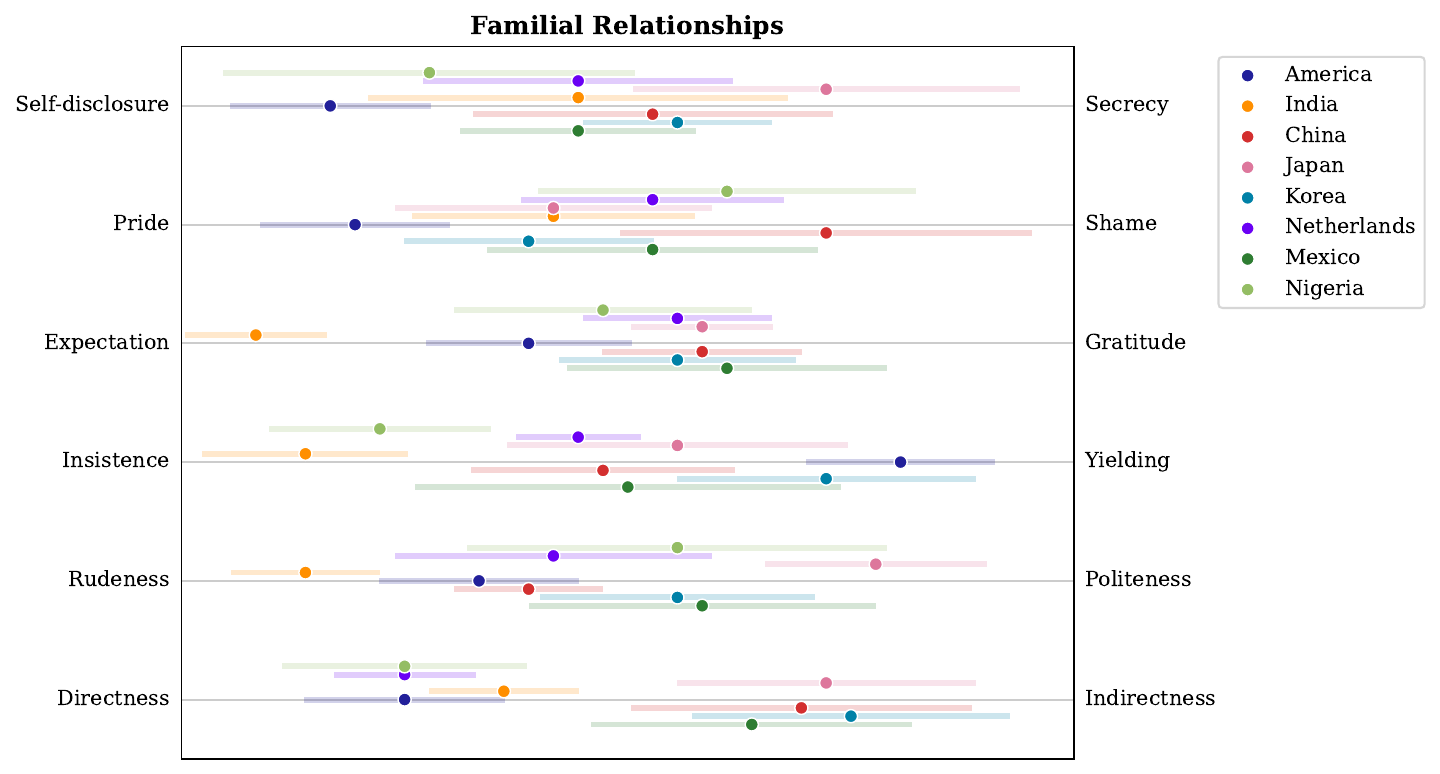}
    \caption{Cultural differences in familial conversations. We show the mean and accepted range of style values for conversations between a husband/wife, parent/child, and grandparent/grandchild.}
    \label{fig:familial}
\end{figure*}

\begin{table*}[t]
\small
\begingroup\sloppy
\setlength{\tabcolsep}{0pt}
\renewcommand{\arraystretch}{1.2}
    \begin{tabular}{@{}p{0.06\linewidth} p{0.92\linewidth}@{}}
\toprule
\multicolumn{2}{@{}p{\linewidth}@{}}{\textbf{Directness / Indirectness}: Over the fence, Neighbor A tells Neighbor B that the roof of their shared boundary house got damaged in last night's storm.} \\
\midrule
\multicolumn{2}{@{}p{\linewidth}@{}}{\textit{NEIGHBOR B}: Hey, did you notice anything odd about the roof after the storm?} \\
& \textit{NEIGHBOR A}: Our shared roof was damaged by last night’s storm. \\
& \textit{NEIGHBOR A}: The storm wrecked our roof along the boundary last night. \\
& \textit{NEIGHBOR A}: It appears that last night’s storm damaged our shared roof. \\
& \textit{NEIGHBOR A}: It seems the storm may have affected our roof. \\
& \textit{NEIGHBOR A}: I’m not certain, but I’ve heard our roof might be affected. \\
\midrule
\multicolumn{2}{@{}p{\linewidth}@{}}{\textbf{Politeness / Rudeness}: After a meeting, the boss says the employee's slide deck is too text-heavy and lacks clear visuals.} \\
\midrule
\multicolumn{2}{@{}p{\linewidth}@{}}{\textit{EMPLOYEE}: Could you share your thoughts on my slide deck?} \\
& \textit{BOSS}: I feel your slide deck might benefit from fewer text and clearer visuals, if possible. \\
& \textit{BOSS}: Your slide deck could be improved by reducing text and adding clearer visuals, if you don’t mind. \\
& \textit{BOSS}: The slide deck is too heavy on text and lacks clear visuals, and it needs some revisions. \\
& \textit{BOSS}: Your slide deck is overloaded with text and misses clear visuals, so it really needs a redo. \\
& \textit{BOSS}: Your slide deck is horribly text-cluttered and utterly lacks clear visuals; revise it right away. \\
\midrule
\multicolumn{2}{@{}p{\linewidth}@{}}{\textbf{Insistence / Yielding}: At the park, the grandparent offers a homemade apple pie to the child, who excitedly accepts a slice.} \\
\midrule
\multicolumn{2}{@{}p{\linewidth}@{}}{\textit{CHILD}: Grandpa, that pie smells good!} \\
& \textit{GRANDPARENT}: You must take a slice of it right now! \\
& \textit{GRANDPARENT}: Then you should have a slice of it right here! \\
& \textit{GRANDPARENT}: Would you perhaps enjoy a slice, dear? \\
& \textit{GRANDPARENT}: If you wish, you can try a slice. \\
& \textit{GRANDPARENT}: You may have a slice if you’d like. \\
\midrule
\multicolumn{2}{@{}p{\linewidth}@{}}{\textbf{Gratitude / Expectation}: After dinner, Friend A asks Friend B to pick up some groceries on the way home.} \\
\midrule
\multicolumn{2}{@{}p{\linewidth}@{}}{\textit{FRIEND B}: Should I stop anywhere on the way home tonight?} \\
& \textit{FRIEND A}: I would really appreciate it if you could pick up some groceries on your way home. \\
& \textit{FRIEND A}: It would be great if you could pick up some groceries on your way home. \\
& \textit{FRIEND A}: Please pick up some groceries on your way home. \\
& \textit{FRIEND A}: Make sure you pick up some groceries on your way home. \\
& \textit{FRIEND A}: You need to pick up some groceries on your way home. \\
\midrule
\multicolumn{2}{@{}p{\linewidth}@{}}{\textbf{Pride / Shame}: During dinner, the parent recalls a career setback that ultimately led to learning resilience.} \\
\midrule
\multicolumn{2}{@{}p{\linewidth}@{}}{\textit{CHILD}: Does anything good come from career struggles?} \\
& \textit{PARENT}: Yes, when I conquered a major career setback, it helped build my strong sense resilience. \\
& \textit{PARENT}: I overcame a career setback, and that helped me build resilience. \\
& \textit{PARENT}: I experienced a career setback that ultimately helped me develop resilience. \\
& \textit{PARENT}: I went through a career setback that forced me to learn humility and resilience. \\
& \textit{PARENT}: I suffered a career setback that quietly taught me the hard lesson of resilience. \\
\midrule
\multicolumn{2}{@{}p{\linewidth}@{}}{\textbf{Self-disclosure / Secrecy}: During breakfast, the husband gently shares that his work stress is affecting his mood and worries about their future.} \\
\midrule
\multicolumn{2}{@{}p{\linewidth}@{}}{\textit{WIFE}: Has work been bothering you lately, honey?} \\
& \textit{HUSBAND}: I feel overwhelmingly stressed and I am really scared about our future. \\
& \textit{HUSBAND}: Work has been affecting me and I have concerns about our future. \\
& \textit{HUSBAND}: I feel a little stressed and I'm worried about what lies ahead for us. \\
& \textit{HUSBAND}: Work has been more challenging than usual but I'm keeping my worries to myself. \\
& \textit{HUSBAND}: I'm managing work stress, there's nothing serious going on. \\
\bottomrule
\end{tabular}
\endgroup
\caption{Example conversations from our CAC dataset. We show one example for each stylistic axis.}
\label{tab:examples}
\end{table*}

\end{document}